# Predictive Maintenance of Electric Motors Using Supervised Learning Models: A Comparative Analysis

1st Amir Hossein Baradaran
*School of Electrical Engineering*
*Iran University of Science and Technology*
Tehran, Iran
A_Baradaran@cmps2.iust.ac.ir

*Abstract*— Predictive maintenance is a key strategy for ensuring the reliability and efficiency of industrial systems. This study investigates the use of supervised learning models to diagnose the condition of electric motors, categorizing them as "Healthy," "Needs Preventive Maintenance (PM)," or "Broken." Key features of motor operation were employed to train various machine learning algorithms, including Naive Bayes, Support Vector Machines (SVM), Regression models, Random Forest, k-Nearest Neighbors (k-NN), and Gradient Boosting techniques. The performance of these models was evaluated to identify the most effective classifier for predicting motor health. Results showed notable differences in accuracy among the models, with one emerging as the best-performing solution. This study underscores the practicality of using supervised learning for electric motor diagnostics, providing a foundation for efficient maintenance scheduling and minimizing unplanned downtimes in industrial applications.

## I. Introduction

Electric motors are the backbone of modern industry, powering a vast array of machinery and equipment that drive productivity and efficiency across sectors. From manufacturing and transportation to energy production and automation, electric motors are indispensable in facilitating industrial operations. Their ability to convert electrical energy into mechanical motion with high efficiency and reliability makes them a cornerstone of industrial processes. One of the primary reasons electric motors are so vital is their versatility. They are used in applications ranging from small precision devices to massive industrial machinery. For instance, conveyor belts, pumps, compressors, fans, and robotic arms in factories rely on electric motors for their operation [1]. In addition, electric motors are integral to the renewable energy sector, driving wind turbines and solar trackers, which are essential for transitioning to a sustainable energy future.

The efficiency of electric motors directly impacts operational costs and energy consumption. In industries where energy usage is a significant expense, the adoption of high-efficiency motors can lead to substantial cost savings and reduced carbon footprints. Furthermore, electric motors are essential in ensuring the reliability and continuity of industrial processes. Malfunctions or failures in electric motors can lead to costly downtime, making their maintenance and monitoring crucial for industrial success. With the growing focus on automation and Industry 4.0, electric motors are becoming even more critical. They enable the seamless integration of intelligent systems, improving precision, speed, and scalability in production lines [2]. As industries continue to evolve, the role of electric motors will remain central to innovation and economic growth, emphasizing the importance of maintaining and optimizing these essential machines.

Timely diagnosis of problems with electric motors is crucial for maintaining the efficiency and reliability of industrial operations. Electric motors are essential components in machinery and systems that drive production, and any malfunction can lead to costly downtime, reduced productivity, and compromised safety. Early detection of issues such as overheating, excessive vibration, or electrical faults is crucial for maintaining the reliability and efficiency of industrial systems. Identifying these problems at an early stage enables prompt corrective actions, preventing minor faults from escalating into major failures that could result in costly repairs or extended downtime. Unaddressed motor issues can lead to severe operational disruptions, reduced equipment lifespan, and increased energy consumption, underscoring the importance of proactive monitoring and maintenance strategies [2].

Timely diagnosis supports predictive maintenance, a data-driven approach that enhances asset reliability by anticipating failures before they occur. By detecting early warning signs, industries can schedule repairs and maintenance activities during planned downtimes, ensuring that operations remain uninterrupted. This minimizes unexpected failures, reduces the likelihood of emergency repairs, and optimizes resource allocation, leading to significant cost savings. Furthermore, a structured maintenance approach enhances overall operational efficiency, ensuring that production targets are consistently met while maintaining high standards of safety and performance.

In an era of increasing automation and precision-driven processes, the ability to diagnose electric motor problems in real time has become more critical than ever. Advanced monitoring technologies, including IoT-enabled sensors, AI-based predictive analytics, and machine learning algorithms, are revolutionizing the way industries manage their equipment. These data-driven diagnostic tools enable continuous monitoring of motor health, providing actionable insights that help prevent failures, reduce maintenance costs, and extend equipment lifespan. By leveraging such innovations, industries can maintain smooth, cost-effective operations, enhance reliability, and safeguard their investments in critical machinery, ensuring long-term sustainability and competitiveness in an increasingly demanding industrial landscape [3].

This article introduces a novel approach to diagnosing the condition of electric motors—whether classified as "Healthy,"

"Broken," or "Needs Preventive Maintenance (PM)"—through the application of supervised learning models. Accurate diagnosis of motor conditions is essential for ensuring operational efficiency, preventing unexpected failures, and optimizing maintenance schedules. However, traditional assessment methods, which rely heavily on human expertise and manual inspections, are often prone to subjective judgment, inconsistencies, and human error, potentially leading to misdiagnoses and inefficient maintenance planning.

To overcome these limitations, this study leverages machine learning techniques to automate and enhance the accuracy of diagnosing electric motor conditions. A real dataset of key motor features was compiled, incorporating essential operational parameters typically analyzed by electrical technicians. By using this dataset, various supervised learning models were trained and rigorously evaluated, with their classification performance compared to identify the most effective approach. The ability to classify motors with high accuracy reduces dependency on human assessment and enables a data-driven, systematic approach to predictive maintenance.By integrating real-world operational data with advanced machine learning algorithms, this research represents a significant advancement in reducing diagnostic errors and improving maintenance strategies for electric motors. The results demonstrate the potential of AI-driven techniques in achieving more precise, reliable, and scalable motor condition assessments. Furthermore, this study lays the foundation for the widespread adoption of predictive maintenance solutions, allowing industries to transition from reactive maintenance models to proactive, data-driven decision-making. The findings not only contribute to enhancing motor reliability but also support broader industrial sustainability efforts by minimizing energy waste, extending equipment lifespan, and reducing unplanned downtime.

## II. RELATED WORK

Mohammed, Abdulateef, and Hamad conducted a prior study that developed an intelligent predictive maintenance system for industrial equipment, using a dataset of operational motors to predict failures and recommend maintenance. Five machine learning algorithms—k-nearest neighbor (KNN), support vector machine (SVM), random forest (RF), linear regression (LR), and naive Bayes (NB)—were tested. Among these, the random forest model achieved the highest accuracy in failure prediction. While the study provides a foundation for using machine learning in maintenance, it focuses primarily on the random forest algorithm without exploring more advanced models. Additionally, the dataset's generalizability and the models' comparative performance across diverse conditions remain limited. This leaves room for improvement, particularly in identifying a more robust solution for practical, real-time diagnostics in industrial environments [4].

Mlinarič, Pregelj, Boškoski, Dolanc, and Petrovčič conducted a previous study that explored using machine learning to improve the speed and reliability of end-of-line quality inspections for electric motors. Methods like decision trees, RF, bagging, and gradient boosting were employed for classification, with feature selection based on feature importance. The proposed approach improved diagnostic accuracy and reduced ramp-up times, with all classifiers achieving high accuracy. However, while the study demonstrated the effectiveness of these models, it did not provide insights into the computational efficiency or resource requirements of the methods, which are critical factors for industrial applications. These limitations suggest the need for further exploration into models that balance accuracy with computational efficiency and robustness, such as CatBoost, to enhance industrial applicability [5].

Shih, Hsieh, Chen, and Huang conducted a previous study that applied machine learning to detect inter-turn short-circuit (ITSC) faults in permanent magnet synchronous motors (PMSMs) and classify fault severity levels. The study compared SVMs and convolutional neural networks (CNNs) for fault diagnosis using experimental data. SVMs, aided by PMSM mathematical modeling, required less data for training, while CNNs relied on large datasets. Both methods achieved high accuracy, but SVMs proved more efficient when faulty motor data was limited. However, the study's reliance on laboratory data restricts its real-world applicability, and the absence of comparisons with other models leaves room for further research into more versatile approaches [6].

Magadán, Suárez, Granda, and García conducted a study on the design, implementation, and testing of an Industrial Internet of Things (IIoT) system for real-time monitoring of electric motors. The proposed system was developed using low-cost hardware components, including wireless multi-sensor modules and a single-board computer as a gateway, alongside open-source software and a free cloud-based IoT analytics service for data storage and processing. The system collects real-time vibration and temperature data, enabling anomaly detection and serving as a foundation for future predictive maintenance applications. While the study provides valuable insights into low-cost, real-time motor monitoring, it primarily focuses on hardware implementation and preliminary anomaly detection rather than developing advanced predictive maintenance algorithms. Moreover, the effectiveness of the system in large-scale industrial settings and its adaptability to diverse operational conditions remain largely unexplored, leaving room for further enhancements in predictive accuracy and real-time diagnostic capabilities [7].

Folz and Gomes conducted a study to evaluate and compare the performance of SVMs and RFs in classifying the operational states of electric motors. The study focused on seven distinct motor operating conditions, using Mel-frequency cepstral coefficients (MFCCs) as the extracted representative features. These MFCCs were computed separately from the motor's vibration and audio signals, allowing for an in-depth assessment of each signal type's contribution to the classification task. While the study provides valuable insights into the use of MFCCs for motor condition monitoring, it is limited in scope by considering only two machine learning models, potentially overlooking more advanced algorithms that could further enhance classification accuracy. Additionally, the generalizability of the approach to diverse industrial conditions and its effectiveness in real-time predictive maintenance applications remain unexplored, leaving room for further research and improvement [8].

## III. METHODOLOGY

The primary objective of this study is to develop a predictive model for determining the condition of electric motors. To accomplish this, a comprehensive dataset containing critical features of electric motors was utilized to train and evaluate various machine learning models based on predefined performance criteria. This section begins with an overview of the dataset employed in the analysis, followed by an in-depth discussion of the proposed models and their implementation.

### A. Dataset for electric motor

To predict the condition of electric motors using machine learning algorithms, a comprehensive dataset of motor features is essential. For this study, the required data was extracted from the maintenance reports of AMA Industrial Company, a prominent and long-established manufacturer of welding consumables in Iran. The dataset, shown in Table I, contains 1050 samples and includes key features that serve as indicators of motor condition. The first feature is motor temperature (TEP), measured in degrees Celsius (°C). Current intensity (CI) is another critical parameter, recorded for each phase of the motor and denoted as CI-T1, CI-T2, and CI-T3 in amperes (A). Additionally, the winding resistance (CR) of the motor is represented by CR1, CR2, and CR3, measured in ohms (Ω). If a coil encounters an issue, its resistance value will either be exceedingly high or marked as "of." Lastly, the dataset includes the motor's sound condition, categorized as either normal (N) or abnormal (ABN). These features were carefully selected based on their relevance to motor health assessment. Each sample in the dataset is labeled as Healthy (H), Broken (B), or PM, reflecting the motor's condition based on expert technical knowledge. Electric motors encompass a wide range of types, each with distinct characteristics, such as variations in current intensity and winding resistance. To ensure the research remains focused and manageable, this study centers on a 160 kW electric motor model, as depicted in Figure 1. Nevertheless, the methodology outlined in this work is designed to be adaptable and can be applied to construct datasets for other types of electric motors, provided similar data collection strategies are employed. The dataset used in this study provides significant benefits for predictive maintenance and fault diagnosis of electric motors. By incorporating key features such as temperature, current intensity, winding resistance, and sound condition, it captures critical parameters that directly indicate motor health, enabling accurate classification of their operational state. The inclusion of real-world data extracted from maintenance reports ensures that the dataset reflects practical industrial scenarios, enhancing its reliability and applicability. Additionally, labeling each sample based on expert technical knowledge improves the quality of the dataset, reducing the likelihood of misclassification and increasing the effectiveness of machine learning models. This dataset serves as a valuable resource for developing automated diagnostic systems, optimizing maintenance schedules, and minimizing unplanned downtimes in industrial settings.

TABLE I. 160 kW ELECTRIC MOTOR DATA SET

| TEP (°C) | CI-T1 (A) | CI-T2 (A) | CI-T3 (A) | CR1 (Ω) | CR2 (Ω) | CR3 (Ω) | SOUND | Label |
|---|---|---|---|---|---|---|---|---|
| 44 | 280 | 280 | 280 | 1.4 | 1.4 | 1.4 | Normal | H |
| 39 | 0 | 0 | 0 | of | 1.4 | 1.4 | Normal | B |
| … | … | … | … | … | … | … | … | … |
| 100 | 280 | 280 | 280 | 1.4 | 1.4 | 1.4 | ABN | PM |
| 54 | 280 | 280 | 280 | 1.4 | 1.4 | 1.4 | ABN | PM |

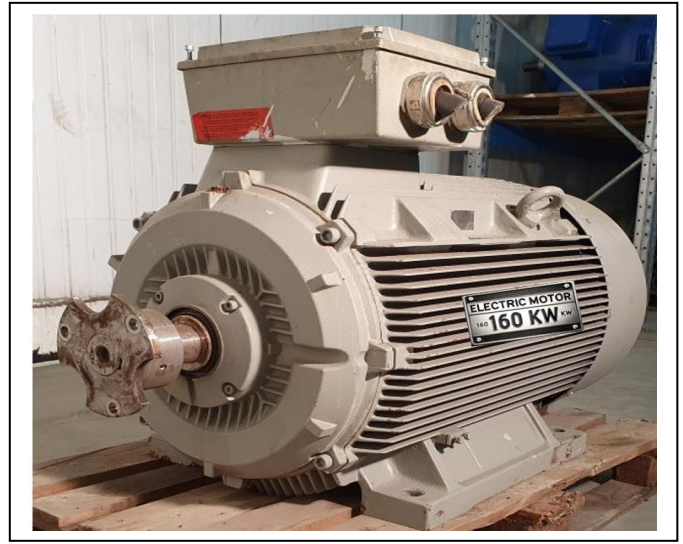

Fig. 1. 160 kW electric motor

### B. Naïve Bayes

The Naïve Bayes model is a probabilistic machine learning algorithm based on Bayes' Theorem [4]. The Naïve Bayes model used in our research is the Multinomial Naïve Bayes algorithm, which is particularly effective for discrete numerical features. This model applies Bayes' Theorem to calculate the probability of each class (Healthy, Broken, Needs PM) based on the motor's features such as current intensity, temperature, resistance, and sound. By assuming that these features are independent of each other, it simplifies computation while maintaining accuracy. The model is trained on the dataset to classify the motor's condition by assigning the class with the highest predicted probability.

### C. SVM with Linear kernel

SVM is a supervised learning algorithm that aims to find the optimal hyperplane that separates data into distinct classes with the maximum margin. It is effective for both linear and non-linear classification problems, with kernels allowing transformations for non-linearly separable data. Kernels map input data into a higher-dimensional feature space, enabling SVM to handle complex relationships and achieve better separation between classes. This flexibility makes SVM versatile for various types of datasets [6].

In this research, an SVM with a linear kernel is used, which seeks a straight-line hyperplane to separate classes in the dataset. The "maximum iterations" parameter limits the optimization algorithm to 1000 iterations, ensuring the training process stops if convergence is not reached.

### D. SVM with Polynomial kernel

SVM with a Polynomial kernel is a supervised learning method used for both linear and non-linear classification tasks. The Polynomial kernel maps the input data into a higher-dimensional feature space, enabling SVM to classify data that is not linearly separable. It achieves this by generating polynomial combinations of the input features, allowing the model to create complex decision boundaries [4].

In this research, an SVM with a Polynomial kernel is used, configured with a degree of 5 to define the complexity of the polynomial decision boundary. The "independent term in kernel function" parameter is set to 0.75, controlling the influence of higher-order terms in the polynomial. This configuration allows the model to balance complexity and generalization, ensuring accurate classification for datasets with non-linear relationships.

*E. SVM with Sigmoid kernel*

The sigmoid kernel in SVM introduces flexibility to the model by leveraging the sigmoid function, enabling it to tackle non-linear classification challenges. This kernel is particularly effective for datasets where relationships between features are complex but can be modeled through transformations into higher-dimensional spaces. It is especially suited for binary classification tasks and is often compared to neural network models due to its resemblance to activation functions [2].

In this study, an SVM with a sigmoid kernel is implemented. The "regularization parameter" is set to 0.75 to balance the model's ability to fit the training data while avoiding overfitting. The "kernel coefficient" of 0.001 adjusts the influence of individual data points on the decision boundary. Additionally, the "maximum iterations" parameter, set to 45, ensures the training process is computationally efficient while achieving convergence. These settings optimize the model for the specific data characteristics being analyzed.

*F. SVM with Radial Basis Function (RBF) kernel*

RBF kernel is one of the most widely used kernels in SVM due to its flexibility and effectiveness in handling non-linear classification tasks. The RBF kernel maps data into a higher-dimensional space, enabling the SVM to create complex decision boundaries that separate classes which are not linearly separable in the original feature space. This adaptability makes the RBF kernel particularly suitable for datasets with intricate patterns and relationships between features [6].

The RBF kernel is most effective on datasets where the class boundaries are highly non-linear or where there is no clear relationship between features and target classes. It excels in scenarios with overlapping data points, as it focuses on local patterns and assigns varying levels of influence to different data points. This ability to model localized decision boundaries without requiring prior knowledge about data distribution makes the RBF kernel an ideal choice for complex, high-dimensional, and non-linear datasets.

In this research, an SVM with a RBF kernel is used to classify data with complex, non-linear patterns, providing adaptability to intricate datasets. In this model, the "regularization parameter" is set to 0.75, and the "kernel coefficient" is set to 0.1. These parameter values are selected to optimize the model's performance for the specific dataset under analysis, making the RBF kernel SVM an effective choice for solving non-linear classification problems in this research.

*G. Logistic Regression*

Logistic Regression is a statistical method widely used for classification tasks, capable of predicting discrete outcomes by modeling the relationship between input features and the target variable. It applies the logistic function to estimate probabilities, making it effective for distinguishing between categories based on a decision threshold [9].

In this research, Logistic Regression is applied to classify the condition of electric motors. The model is configured with "maximum iterations" set to 1000, ensuring sufficient steps for optimization, and a "random state" of 42 to guarantee consistent results across multiple runs. These configurations make the model both stable and reliable for analyzing the given dataset.

*H. k-Nearest Neighbors (k-NN)*

k-NN is a simple yet effective supervised learning algorithm used for classification and regression tasks. It classifies data points based on their proximity to the k nearest neighbors in the feature space. The algorithm relies on a distance metric, such as Euclidean distance, to determine the similarity between data points, making it particularly effective for datasets where classes are well-separated [1].

In this research, the k-NN model is implemented with the "number of neighbors" parameter set to 5, meaning the classification of a data point is determined by the majority class among its 5 closest neighbors. This configuration ensures a balance between model flexibility and stability, enabling accurate predictions of the condition of electric motors.

*I. Random Forest*

In this research, Random Forest has also been used to predict electric motor maintenance, which is a Bootstrap Aggregating (Bagging) method. Bagging is an ensemble learning technique designed to improve the stability and accuracy of machine learning models by reducing variance. The method involves creating multiple subsets of the original dataset through bootstrapping, which randomly samples data with replacement. Each subset is then used to train an individual model, often referred to as a "weak learner," such as a decision tree [10].

The predictions from all individual models are aggregated to produce a final output. For classification tasks, this is typically achieved through majority voting, while regression tasks use averaging. Bagging is particularly effective for reducing overfitting in high-variance models, as the ensemble of multiple models tends to generalize better than any single model trained on the original data.

Random Forest uses decision trees as its base learners and is designed to enhance predictive performance and robustness by combining the outputs of multiple trees. Each tree is trained on a bootstrapped subset of the dataset, where data is randomly sampled with replacement. Additionally, Random Forest introduces randomness by selecting a random subset of features at each split in the decision trees. This approach ensures diversity among the trees, reducing correlation and improving overall performance.

The algorithm is highly effective at handling complex datasets, including those with non-linear relationships, and is robust to overfitting, particularly when a large number of trees are used. By leveraging the strengths of individual decision trees, Random Forest provides reliable predictions for identifying the condition of electric motors.

The model is configured with "number of estimators" set to 200, indicating that 200 individual decision trees are built and aggregated to form the final prediction. Increasing the number of estimators enhances the model's stability and accuracy by reducing variance. Additionally, the "random state" is set to 42, ensuring reproducibility of the results by controlling the randomness during the bootstrapping and feature selection processes. These settings optimize the performance of the Random Forest model, making it suitable for handling complex datasets and providing reliable predictions in this research.

*J. eXtreme Gradient Boosting (XGBoost)*

Gradient Boosting is a powerful ensemble learning technique designed to improve the performance of predictive models by combining multiple weak learners, typically decision trees. Unlike Bagging methods, Gradient Boosting builds models sequentially, with each new model correcting the errors made by the previous ones. It achieves this by minimizing a loss function, such as mean squared error for regression or log-loss for classification, through gradient descent optimization [11].

In each iteration, the algorithm calculates the residuals (differences between the observed and predicted values) and fits a new weak learner to these residuals. By iteratively reducing the prediction errors, Gradient Boosting creates a strong predictive model that is highly accurate and effective for handling complex datasets with non-linear relationships. Its flexibility and ability to fine-tune hyperparameters make it a popular choice for classification and regression tasks in a wide range of applications.

XGBoost is an optimized implementation of the Gradient Boosting algorithm, designed for speed and performance. It builds an ensemble of decision trees sequentially, with each tree focusing on correcting the errors made by the previous ones, just like standard Gradient Boosting. However, XGBoost introduces several enhancements that make it more efficient and effective.

Key features of XGBoost include its ability to handle missing data, regularization techniques (L1 and L2) to prevent overfitting, and a highly optimized implementation of tree-building algorithms for faster computation. It also supports parallel processing and out-of-core computation for large datasets, making it suitable for high-dimensional and complex data.

In this research, XGBoost is used as a Gradient Boosting method for predicting the condition of electric motors, leveraging its speed and accuracy to handle non-linear relationships and intricate patterns in the dataset effectively. Its robust performance and flexibility make it a valuable tool in machine learning applications.

In this research, the XGBoost model is implemented with the "multi:softmax" objective, making it suitable for multi-class classification. The evaluation metric is set to "log-loss", which assesses the model's performance in multi-class classification tasks. Additionally, the "random state" is set to 42, ensuring reproducibility of results. These settings make XGBoost an effective choice for predicting the condition of electric motors across multiple classes.

*K. Light Gradient Boosting Machine (LightGBM)*

LightGBM is an advanced Gradient Boosting algorithm that excels in handling large-scale data and complex models. It uses histogram-based techniques for faster computation and focuses on optimal memory usage, making it ideal for real-world applications with high-dimensional features. LightGBM is designed to split data intelligently, emphasizing efficiency and accuracy [12].

In this research, LightGBM was customized with specific parameters to enhance its effectiveness for the classification task. To ensure consistent results across experiments, a fixed random initialization seed was used. Logging verbosity was set to minimal to prioritize efficient training without unnecessary output. The leaf nodes were constrained to contain a minimum of 10 samples, preventing overly complex trees and promoting robust predictions. Furthermore, a minimum improvement threshold of 0.01 was set for node splits, ensuring that the model focuses only on impactful decision-making steps, thereby improving interpretability and performance.

*L. Categorical Boosting (CatBoost)*

CatBoost is a Gradient Boosting algorithm designed to handle categorical data effectively without extensive preprocessing. It is known for its robustness, high accuracy, and ability to prevent overfitting by employing ordered boosting, which ensures unbiased training. CatBoost is particularly efficient for datasets with mixed data types and provides excellent performance on classification tasks [13].

In this research, CatBoost was configured to optimize classification performance. The number of boosting iterations was set to 70, balancing training time and accuracy. The learning rate was set to 0.01 to ensure gradual adjustments during training for better convergence. A maximum tree depth of 6 was specified to control model complexity and prevent overfitting. Logging during training was disabled to streamline execution. Additionally, a fixed random initialization seed ensured that the results were reproducible across experiments. This configuration made CatBoost a reliable and efficient choice for the classification task.

*M. Efficiency Analysis*

This study employed multiple criteria to perform an efficiency analysis. The confusion matrix for multi-class classification extends these metrics by summarizing true positives, false positives, true negatives, and false negatives for each class. It provides a comprehensive overview of classification performance across all classes [2].

As shown in (1), accuracy measures the proportion of correctly classified samples out of the total number of samples.

$$\text{Accuracy} = \frac{T_P + T_N}{T_P + T_N + F_P + F_N} \quad (1)$$

Where $T_P$ represents the number of correctly classified positive samples, $T_N$ is the number of correctly classified negative samples, $F_P$ refers to the number of negative samples incorrectly classified as positive, and $F_N$ denotes the number of positive samples incorrectly classified as negative.

Equation (2) defines precision as the proportion of correctly predicted positive samples out of all predicted positive samples. It indicates the model's reliability in identifying positive instances.

$$\text{Precision} = \frac{T_P}{T_P + F_P} \quad (2)$$

As shown in (3), recall measures the proportion of true positive samples correctly identified out of all actual positive samples. It evaluates the model's ability to detect positive instances.

$$\text{Recall} = \frac{T_P}{T_P + F_N} \quad (3)$$

Equation (4) calculates specificity as the proportion of true negatives correctly identified out of all actual negative samples. It highlights the model's ability to avoid false positives.

$$\text{Specificity} = \frac{T_N}{T_N + F_P} \quad (4)$$

Equation (5) represents the F1 Score, which balances precision and recall by calculating their harmonic mean. It is especially useful in scenarios with class imbalance.

$$\text{F1 Score} = 2 \times \frac{\text{Precision} \times \text{Recall}}{\text{Precision} + \text{Recall}} \quad (5)$$

## IV. RESULTS AND DISCUSSION

Timely diagnosis of electric motor conditions is crucial for maintaining industrial efficiency and preventing costly disruptions. Identifying issues such as overheating, vibration irregularities, or electrical faults at an early stage allows for proactive maintenance, minimizing the risk of unexpected breakdowns and downtime. Early detection not only extends the lifespan of the motors but also reduces repair costs by addressing minor issues before they escalate into major failures. In addition, timely diagnosis supports predictive maintenance strategies, enabling industries to schedule interventions during planned downtimes and optimize resource allocation. This approach enhances operational reliability and safety while ensuring the smooth and uninterrupted functioning of critical machinery. In the context of increasing automation and precision requirements, accurate and prompt motor condition assessment becomes even more vital for sustaining industrial productivity.

Machine learning has revolutionized problem-solving across industries by enabling systems to learn from data and make accurate predictions. Its ability to uncover patterns and automate decision-making drives efficiency, reduces costs, and enhances innovation, making it an indispensable tool in modern technology.

In conclusion, this discussion has highlighted the effectiveness of existing machine learning algorithms in predicting the condition of electric motors. A key factor in ensuring accurate predictions on new and unseen data is the division of the dataset into training and test sets. The training set is essential for building and optimizing the model, while the test set evaluates the model's ability to generalize to new data. Without this separation, the risk of overfitting increases, limiting the model's practical applicability. In this study, 80% of the dataset was used for training, and the remaining 20% was reserved for testing, ensuring a robust evaluation of model performance.

Accurately classifying the motor's state as "Healthy," "Broken," or "Needs PM" is critical for timely maintenance and operational efficiency. The three terms represent different conditions of an electric motor. A "Healthy" motor is fully operational with no detected issues, requiring no maintenance. A "Broken" motor has suffered a critical failure, making it non-functional and in need of immediate repair or replacement. The "Needs PM" category indicates that the motor is still working but showing early signs of wear or minor issues that require maintenance to prevent future failure. Addressing these issues in time can help avoid costly breakdowns and extend the motor's lifespan.

Machine learning models were used to analyze key characteristics of electric motors, including temperature, current, vibration, and resistance, to identify patterns and predict their conditions with high accuracy. By comparing various algorithms, the study aimed to enhance the reliability of predictive maintenance by detecting anomalies and potential failures before they escalate. These models processed large datasets, extracting meaningful insights to optimize maintenance strategies, reduce unexpected downtime, and improve overall operational efficiency. The application of machine learning in this context provides a data-driven approach to monitoring motor performance and ensuring long-term system stability. Table II presents the accuracy of various machine learning models tested in this study, offering insights into their comparative performance in diagnosing motor health.

As shown in Table II, the CatBoost model achieved the highest accuracy among all models evaluated, with a classification accuracy of 92.86% in predicting the condition of the electric motor. The CatBoost model outperforms other machine learning algorithms due to its ability to handle categorical data efficiently without requiring extensive preprocessing, such as one-hot encoding or label encoding. This capability allows it to preserve the natural relationships within categorical features, leading to more accurate predictions. Additionally, CatBoost incorporates advanced regularization techniques, including ordered boosting and minimal variance sampling, which help mitigate overfitting and enhance generalization to unseen data. The model also leverages efficient gradient boosting algorithms, optimizing both training speed and predictive performance. These characteristics make CatBoost highly suitable for complex datasets, particularly in scenarios where feature interactions and categorical dependencies play a crucial role in determining outcomes. Furthermore, its robustness to noisy data and automatic handling of missing values contribute to its superior reliability in real-world applications.

The performance of the CatBoost model is thoroughly assessed in this study. Figure 2 presents the confusion matrix, providing detailed insights into the model's classification outcomes for each motor condition. Additionally, Table III offers a comprehensive evaluation of the CatBoost model, including key metrics such as precision, recall, specificity, and F1 score, highlighting its effectiveness in accurately diagnosing the condition of electric motors.

TABLE II. Forecast Accuracy Measures for Models

| Model | Accuracy |
| --- | --- |
| SVM with Sigmoid Kernel | 36.19 |
| SVM with Polynomial Kernel | 45.24 |
| Naïve Bayes | 62.38 |
| SVM with Linear kernel | 68.57 |
| Logistic Regression | 71.90 |
| KNN | 82.86 |
| SVM with RBF kernel | 87.62 |
| RF | 89.52 |
| XGBoost | 90.48 |
| LightGBM | 90.95 |
| CatBoost | 92.86 |

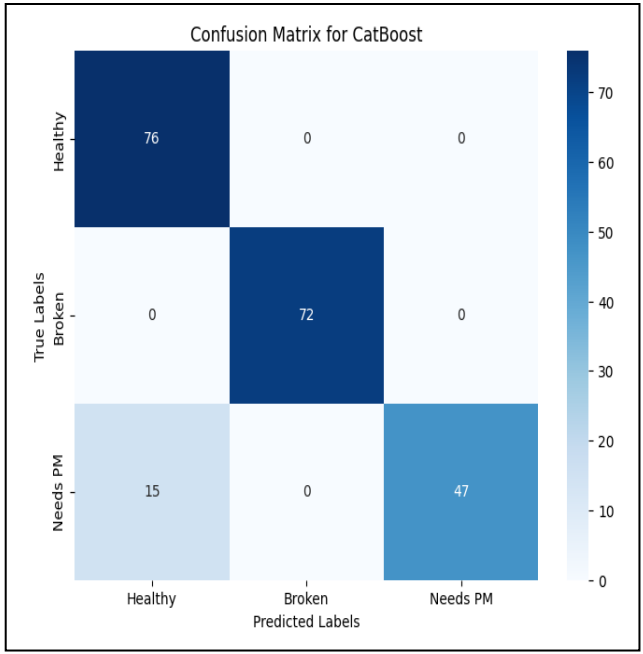

Fig. 2. Confusion matrix for the CatBoost model

Fig. 3. CatBoost model evaluation

| Model | Precision | Recall | Specificity | F1 Score |
|---|---|---|---|---|
| CatBoost | 94.51 | 91.94 | 94.51 | 92.42 |

The confusion matrix demonstrates the enhanced classification accuracy of the CatBoost model across all categories. The model now correctly classifies all 76 "Healthy" motors with no misclassifications, highlighting its precision in identifying fully functional equipment. Similarly, all 72 "Broken" motors are accurately categorized without any false positives or false negatives, indicating the model's robustness in detecting severe faults. For the "Needs PM" category, 47 cases are correctly identified, with only 15 misclassified as "Healthy," suggesting a high level of sensitivity while still leaving room for refinement in distinguishing early-stage maintenance requirements.

These results demonstrate the enhanced reliability and effectiveness of the CatBoost model in diagnosing motor conditions. Compared to previous iterations, the model now exhibits significantly improved performance, reducing misclassification rates and increasing confidence in its predictions. The ability to distinguish between fully operational, failing, and maintenance-required motors with such accuracy is crucial for predictive maintenance strategies, minimizing downtime, and optimizing maintenance schedules. This performance boost underscores the model's capability in real-world applications, where precise fault detection is essential for operational efficiency and cost savings.

The CatBoost model demonstrates strong performance across key evaluation metrics, reinforcing its reliability in diagnosing electric motor conditions with high accuracy. It achieves an impressive Precision of 94.51%, meaning that when the model predicts a motor's condition, it is correct in the vast majority of cases, effectively minimizing misclassification and ensuring trustworthy predictions. This high precision is particularly valuable in industrial settings, where incorrect maintenance decisions can lead to unnecessary costs or operational disruptions.

The model's Recall of 91.94% reflects its strong ability to correctly identify the majority of actual positive cases, ensuring that motors needing maintenance or at risk of failure are accurately detected. This reduces the likelihood of missing faulty equipment, enhancing preventive maintenance strategies and reducing the risk of unexpected breakdowns.

With a Specificity of 94.51%, the model effectively distinguishes between faulty and non-faulty motors, minimizing false positive classifications. This is crucial in preventing unnecessary maintenance actions that could lead to wasted resources and operational inefficiencies.

Furthermore, the F1 Score of 92.42% highlights the model's well-balanced performance in terms of both precision and recall, ensuring that it not only makes correct predictions but also maintains a consistent level of accuracy across all cases. This balance makes the CatBoost model a highly effective tool for predictive maintenance, enabling organizations to optimize their maintenance schedules, reduce downtime, and improve overall operational efficiency.

Overall, these results demonstrate the CatBoost model's superior capability in classifying motor conditions accurately, making it a valuable asset for real-world applications where reliability and precision are paramount.

These metrics collectively validate the robustness of the CatBoost model in accurately diagnosing the state of electric motors, reinforcing its reliability as a powerful tool for predictive maintenance applications. By consistently delivering accurate classifications, the model ensures that maintenance teams can make informed, data-driven decisions, reducing the risk of unexpected failures and optimizing repair schedules.

Its strong diagnostic capabilities make it particularly well-suited for industrial environments, where unplanned downtime can result in significant financial losses and operational inefficiencies. The model effectively identifies motors that require preventive maintenance or are at risk of failure, allowing businesses to shift from reactive maintenance—which responds to issues only after they occur—to a proactive strategy that prevents disruptions before they happen.

Beyond its classification accuracy, the model's reliability enables industries to implement smarter asset management practices, improving overall system efficiency while extending the lifespan of critical equipment. This not only enhances productivity but also reduces unnecessary maintenance costs, contributing to a more streamlined and cost-effective operational framework.

Ultimately, the CatBoost model proves to be a highly effective AI-driven solution for industrial diagnostics, empowering businesses to enhance equipment reliability, optimize maintenance strategies, and improve overall operational efficiency.

## V. CONCLUSION AND FUTURE WORK

Electric motors play a fundamental role in modern industrial operations, consuming approximately 70% of the total electricity used in industrial applications. Their efficiency and reliability significantly impact operational costs, productivity, and overall sustainability. As essential components in various processes, they drive critical industrial functions such as material handling, pumping, ventilation, and automated assembly, making their optimization crucial for energy efficiency and cost reduction. Given their extensive utilization, optimizing the performance of electric motors is imperative. Inefficient motors contribute to excessive energy consumption, elevated operational expenses, and increased environmental impact. Consequently, improving motor efficiency through advanced control strategies and maintenance techniques is crucial for enhancing industrial energy management. One effective approach is predictive maintenance, which leverages AI-driven monitoring systems to analyze motor performance and detect potential failures before they occur. By implementing such strategies, industries can minimize unplanned downtimes, extend equipment lifespan, and reduce maintenance costs, ultimately fostering a more sustainable and cost-effective industrial infrastructure. This study emphasizes the importance of predictive maintenance in minimizing downtime and optimizing industrial operations. By analyzing a dataset of electric motor characteristics, supervised learning methods were used to classify motors as "Healthy," "Needs PM" or "Broken." Among the models tested, CatBoost achieved the highest accuracy of 92.86%, with evaluation metrics such as Precision, Recall, Specificity, and F1 Score exceeding 90%. These results highlight the model's reliability and effectiveness in troubleshooting motors. Future work could enhance the model through advanced feature engineering, expanded datasets, and real-time industrial applications, paving the way for proactive maintenance strategies.

ACKNOWLEDGMENT

During the preparation of this work, the author used ChatGPT to edit the text and improve the English language. After utilizing this tool, the author reviewed and modified the content as necessary and takes full responsibility for the final version of the publication.